\documentclass[11pt]{article}

\usepackage[final]{acl}

\usepackage{times}
\usepackage{latexsym}

\usepackage[T1]{fontenc}

\usepackage[utf8]{inputenc}

\usepackage{microtype}

\usepackage{inconsolata}

\usepackage{graphicx}

\usepackage{booktabs}
\usepackage{multirow}
\usepackage{makecell}
\usepackage{colortbl}
\usepackage{amssymb}
\usepackage{amsmath}
\usepackage[table]{xcolor}
\usepackage{colortbl}
\definecolor{lightred}{RGB}{255,230,230}

\title{Why Does Reinforcement Learning Generalize?
A Feature-Level Mechanistic Study of Post-Training in Large Language Models}

\author{
\textbf{Dan Shi}$^{1}$, \textbf{Zhuowen Han}$^{1}$, \textbf{Simon Ostermann}$^{2,3}$, \textbf{Renren Jin}$^{1}$, \\
\textbf{Josef van Genabith}$^{2,3}$, \textbf{Deyi Xiong}$^{1}$\thanks{Corresponding author} \\
$^{1}$TJUNLP Lab, School of Computer Science and Technology, Tianjin University, China \\
$^{2}$German Research Center for Artificial Intelligence (DFKI), Saarbrücken, Germany \\
$^{3}$Saarland University, Saarbrücken, Germany \\
\texttt{\{shidan, dyxiong\}@tju.edu.cn}
}

\begin{document}
\maketitle

\begin{abstract}
Reinforcement learning (RL)-based post-training often improves the reasoning performance of large language models (LLMs) beyond the training domain, while supervised fine-tuning (SFT) frequently leads to general capabilities forgetting. However, the mechanisms underlying this contrast remain unclear.
To bridge this gap, we present a feature-level mechanistic analysis methodology to probe RL generalization using a controlled experimental setup, where RL- and SFT-tuned models are trained from the same base model on identical data. Leveraging our interpretability framework, we align internal activations across models within a shared feature space and analyze how features evolve during post-training.
We find that SFT rapidly introduces many highly specialized features that stabilize early in training, whereas RL induces more restrained and continually evolving feature changes that largely preserve base models' representations. Focusing on samples where RL succeeds but the base model fails, we identify a compact, task-agnostic set of features that directly mediate generalization across diverse tasks. Feature-level interventions confirm their causal role: disabling these features significantly degrades RL models' generalization performance, while amplifying them improves base models' performance. The code is available at \url{https://github.com/danshi777/RL-generalization}.

\end{abstract}

\section{Introduction}

Reinforcement learning (RL) has emerged as a powerful paradigm for enhancing the reasoning capabilities of large language models (LLMs), particularly in solving complex logical tasks involving mathematics and programming \cite{guo2025deepseek, jaech2024openai, team2025kimi}. Notably, models optimized via RL on narrowly defined reasoning objectives often demonstrate substantial performance improvements on tasks far beyond their training distribution. In contrast, supervised fine-tuning (SFT) is frequently observed to induce degradation or forgetting of previously acquired general-purpose capabilities \cite{huan2025does, chu2025sft}.

Despite these consistent empirical findings, why RL-tuned models can generalize well remains poorly understood. Unlike SFT, which distills full reasoning trajectories from a teacher model, RL typically relies only on outcome-level supervision. From a mechanistic perspective, it is unclear how such weak and indirect signals yield broad, transferable improvements across diverse tasks.

In this work, we address this gap through a feature-level interpretability framework designed to investigate how SFT and RL differentially reshape internal representations. We first employ Sparse Crosscoder to align the internal activations of the base model with those of its RL- and SFT-tuned counterparts within a shared, interpretable feature space. This alignment allows us to systematically compare internal representations across models and to track how individual features emerge, evolve, and diverge during post-training. 

However, pairwise comparisons alone are insufficient for fully characterizing the relationship between SFT and RL representations. To overcome this limitation, we then propose a three-model Sparse Crosscoder that jointly aligns the base, SFT-trained, and RL-trained models within a single sparse feature space, and a novel Model Attribution Score (MAS) to measure feature specificity. This unified representation enables direct attribution of each feature to a specific training paradigm.

Using this framework, we conduct a systematic analysis of feature dynamics throughout training. Our results reveal a clear and consistent distinction between the two paradigms. SFT rapidly introduces a large number of highly specialized features that stabilize early in training, whereas RL induces more restrained and continuously evolving feature changes, largely preserving the base model’s representational structure. This difference provides an initial representational explanation for why SFT tends to memorize task-specific patterns, while RL maintains broader capabilities.

Beyond descriptive analysis, our framework enables direct mechanistic investigation of RL-induced generalization. We identify a compact set of internal features that actively control cross-task transfer by focusing on samples where generalization explicitly occurs, cases in which the RL-tuned model succeeds while the base model fails, and measuring feature-level activation differences within the aligned feature space.

Through targeted feature interventions, we further demonstrate that these features are causally responsible for generalization: disabling them in RL-tuned models leads to significant performance degradation, while amplifying them in base models induces substantial gains, even on unseen tasks. These results indicate that RL does not merely improve task-specific performance, but instead strengthens a compact, task-agnostic set of features that governs generalization behavior.

In summary, our contributions are as follows:

\begin{itemize}
    \item We propose a feature-level interpretability framework for mechanistically analyzing how different post-training paradigms reshape internal representations and give rise to RL generalization.

    \item We present a three-model Sparse Crosscoder and a Model Attribution Score (MAS) that enable unified feature alignment and attribution across base, SFT, and RL models.

    \item We reveal a fundamental distinction between post-training paradigms: SFT induces early-stabilized, highly specialized features, whereas RL yields more restrained and continuously evolving feature changes, and we identify a small set of generalization-controlling features that provide a direct mechanistic explanation for RL generalization.
\end{itemize}

\section{Related Work}

\paragraph{Feature-Level Interpretability in LLMs.} As LLMs continue to exhibit diverse and sophisticated capabilities \cite{guo2023evaluating, chang2024survey, shi2024corecode}, understanding the internal mechanisms that give rise to these behaviors has emerged as a prominent direction \cite{DBLP:conf/acl/DaiDHSCW22, shi2024ircan, DBLP:conf/acl/ChenHFL24, DBLP:conf/emnlp/WuLXDW0X23, shi2024large}. Recent work has explored feature-level interpretability in LLMs, often using Sparse Autoencoders (SAEs, \citealp{ cunningham2023sparse, gao2024scaling, markssparse}) to identify internal features associated with specific functions, such as emotions \cite{han-etal-2025-towards}, safety behaviors \cite{weng2025safe, yeon2025gsae}, language-specific representations \cite{deng2025unveiling}, and reasoning processes \cite{galichin2025have}. Sparse Crosscoders extend SAEs, have further enabled comparisons between fine-tuned and base models, revealing features introduced by SFT \cite{lindsey2024sparsecrosscoders, baek2025towards}. In contrast, our work compares both SFT- and RL-trained models against the same base model and introduces a three-model sparse crosscoder to jointly analyze their feature differences.

\paragraph{Generalization Comparison of SFT and RL.} Several studies have compared the generalization of SFT- and RL-tuned models at the behavioral level, documenting performance gains and losses across diverse tasks. These works consistently report that RL-tuned models generalize more robustly across domains, while SFT often leads to forgetting of general capabilities \cite{huan2025does, chu2025sft}. However, the internal mechanisms underlying this contrast remain largely unexplored. Although \citet{huan2025does} attempt to explain this difference by relating it to larger representation or output distribution drift induced by SFT, their analysis remains at the level of global representations and does not identify the specific internal mechanisms that causally drive generalization. Our work addresses this gap by providing a feature-level, causally validated explanation of RL-induced generalization.

\section{Interpretability Methodology}

To systematically analyze how different post-training paradigms reshape internal representations, we introduce a unified feature-level interpretability framework. The framework is guided by three core principles. First, the models under comparison must be strictly comparable, such that observed representational differences can be attributed solely to the training paradigm. Second, internal representations from different models must be aligned into a shared feature space to enable direct, fine-grained comparison of features. Third, the framework should not only support descriptive analysis of feature differences, but also provide direct evidence for the generalization capacity of RL.

To satisfy these principles, we design three key components in our interpretability framework: (i) sparse crosscoders for feature-level alignment and attribution between the base model and tuned models, (ii) a three-model extension that enables joint comparison across base, SFT, and RL models, and (iii) a method for identifying generalization-controlling features. We introduce each component in turn below.

\subsection{Feature-Level Alignment and Attribution via Sparse Crosscoders}

A central challenge in comparing internal representations across independently trained models is that their activation spaces are not directly aligned. To address this challenge, we employ Sparse Crosscoders \citep{lindsey2024sparsecrosscoders} to align activations from the different models into a shared sparse feature space, enabling direct, meaningful comparison of representational changes induced by tuning. 

\paragraph{Two-Model Sparse Crosscoder for Comparison of Tuned Models and Base Model.}
Sparse Crosscoders extend SAEs by jointly encoding activations from different sources, such as different models, layers, or positions, into a shared sparse feature space. This formulation enables a fine-grained comparison of representational changes induced by tuning.

Formally, the encoder computes feature activations as:
\begin{equation}
\boldsymbol{f}(x_j)
= \mathrm{ReLU}\left(
\sum_{i \in {O, T}}
\boldsymbol{W}^{(i)}_{\mathrm{enc}} \boldsymbol{a}^{(i)}(x_j)+\boldsymbol{b}_{\mathrm{enc}}
\right),
\end{equation}
and the decoder reconstructs the activations for each model:
\begin{equation}
    \boldsymbol{\hat{a}}^{(i)}(x_j) = \boldsymbol{W}^{(i)}_{\mathrm{dec}} \, \boldsymbol{f}(x_j) + \boldsymbol{b}^{(i)}_{\mathrm{dec}}.
\end{equation}
Here, $O$ denotes the original base model and $T$ represents the tuned model (either SFT or RL). $\boldsymbol{a}^{(i)}(x_j)\in \mathbb{R}^{d_\mathrm{model}}$ is the residual-stream activation of model $i$ at token $x_j$, $\boldsymbol{f}(x_j)\in \mathbb{R}^{d_\mathrm{sparse}}$ is the sparse feature activations, and $\boldsymbol{\hat{a}}^{(i)}(x_j)$ is the reconstructed activation. 
The crosscoder is trained by minimizing a reconstruction loss $\mathcal{L}_{\mathrm{recon}}$ with sparsity regularization $\mathcal{L}_{\mathrm{sparsity}}$:
\begin{equation}
\begin{aligned}
\mathcal{L}
&= \mathcal{L}_{\mathrm{recon}} + \beta \mathcal{L}_{\mathrm{sparsity}}, \\
\mathcal{L}_{\mathrm{recon}}
&= \sum_{i = O,T} \left\lVert \boldsymbol{a}^{(i)} - \boldsymbol{\hat{a}}^{(i)} \right\rVert^{2}, \\
\mathcal{L}_{\mathrm{sparsity}}
&= \sum_{k} f_k(x_j) \sum_{i = O,T}
\left\lVert \boldsymbol{W}^{(i)}_{\mathrm{dec},k} \right\rVert.
\end{aligned}
\end{equation}
This objective forces the model to learn a small set of interpretable features that capture the distinct properties of the activations. Within our framework, this two-model sparse crosscoder serves as the basic building block for pairwise comparison between a tuned model and its base counterpart.

\paragraph{Identifying Model-Specific Features with NRN}

To quantify how strongly each feature is unique to each model, we follow \citet{baek2025towards} using the Normalized Relative Norm (NRN) metric to analyze the features. NRN is computed as the ratio between the L1 norm of the decoder vector for each model:
\begin{equation}
    \mathrm{NRN}
    = \frac{\left\lVert \boldsymbol{W}^{(T)}_{\mathrm{dec},k} \right\rVert_{1}}
           {\left\lVert \boldsymbol{W}^{(O)}_{\mathrm{dec},k} \right\rVert_{1}+\left\lVert \boldsymbol{W}^{(T)}_{\mathrm{dec},k} \right\rVert_{1}},
\end{equation}
where $O$ denotes the original base model, $T$ represents the tuned model (SFT or RL), and $\boldsymbol{W}^{(\cdot)}_{\mathrm{dec}}$ denotes the corresponding crosscoder's decoder. 
Intuitively, if a feature contributes more to reconstructing the tuned model’s activations, the decoder assigns it a larger norm in $\boldsymbol{W}^{(T)}_{\mathrm{dec}}$, yielding NRN → 1. In this case, NRN → 1 corresponds to features that are unique to the tuned model, NRN → 0 corresponds to features unique to the original model, while NRN = 0.5 corresponds to shared features.

\paragraph{Three-Model Sparse Crosscoder for Joint Representation Comparison.}

The two-model Sparse Crosscoder enables fine-grained comparisons between a tuned model and its base counterpart. However, such pairwise analyses are inherently limited when multiple training paradigms are involved. In particular, when comparing an SFT-tuned model and an RL-tuned model against the same base model, pairwise crosscoders cannot disentangle whether a feature is shared by all models, specific to one tuned model, or shared by the two tuned models but absent in base model. 

More importantly, features identified in different pairwise crosscoders are not directly comparable. For example, a feature with a given index (e.g., feature \#2026) in the SFT-Base crosscoder does not necessarily correspond to the feature with the same index in the RL-Base crosscoder, since the two crosscoders are trained independently and may learn entirely different sparse bases.

To address this limitation, we introduce a \textbf{three-model Sparse Crosscoder}, which jointly aligns the base, SFT-tuned, and RL-tuned models within a single sparse feature space. By unifying all three models within a single sparse basis, the three-model sparse crosscoder constitutes a key component of our interpretability framework, enabling direct, simultaneous comparison of feature sharing and specialization across post-training paradigms.

Specifically, given a token $x_j$, we jointly encode the residual-stream activations from the three models into a shared sparse feature representation: 
\begin{equation}
\boldsymbol{f}(x_j) = \mathrm{ReLU}\!\left( \sum_{i=O,S,R} \boldsymbol{W}_{\mathrm{enc}}^{(i)} \boldsymbol{a}^{(i)}(x_j) + \boldsymbol{b}_{\mathrm{enc}} \right),
\end{equation}
where $O$, $S$, and $R$ denote the original base model, the SFT-tuned model, and the RL-tuned model, respectively. $\boldsymbol{a}^{(i)}(x_j)$ represents the activation of model $i$, and $\boldsymbol{f}(x_j)$ is the shared sparse feature vector. Then, the decoder reconstructs the activations for each model:
\begin{equation}
\boldsymbol{\hat{a}}^{(i)}(x_j) = \boldsymbol{W}_{\mathrm{dec}}^{(i)} \boldsymbol{f}(x_j) + \boldsymbol{b}_{\mathrm{dec}}^{(i)}.
\end{equation}

The training objective minimizes a combination of reconstruction loss and sparsity regularization:
\begin{equation}
\begin{aligned}
\mathcal{L}
&= \sum_{i=O,S,R}
   \left\| \boldsymbol{\hat{a}}^{(i)} - \boldsymbol{a}^{(i)} \right\|^{2} \\
&\quad + \sum_{k} f_k(x_j)
   \sum_{i=O,S,R}
   \left\| \boldsymbol{W}_{\mathrm{dec},k}^{(i)} \right\|.
\end{aligned}
\end{equation}

This objective encourages the model to discover a compact set of sparse features that jointly explain the activations of all three models, while allowing each feature to contribute unequally to different models.

\paragraph{Measuring Feature Specificity with MAS.}
 
To quantify how strongly each feature is attributed to a particular model within the shared crosscoder space, we define a three-way normalization, which we refer to as the Model Attribution Score (MAS). For each feature $k$, the formula for calculating MAS is as follows:
\vspace{-1mm}
\begin{equation}
    \mathrm{MAS}_{O}
\resizebox{0.73\linewidth}{!}{$
    = \frac{\left\lVert \boldsymbol{W}^{(O)}_{\mathrm{dec},k} \right\rVert_{1}}
           {\left\lVert \boldsymbol{W}^{(O)}_{\mathrm{dec},k} \right\rVert_{1}+\left\lVert \boldsymbol{W}^{(S)}_{\mathrm{dec},k} \right\rVert_{1}+\left\lVert \boldsymbol{W}^{(R)}_{\mathrm{dec},k} \right\rVert_{1}}
$},
\end{equation}
\begin{equation}
    \mathrm{MAS}_{S}
\resizebox{0.73\linewidth}{!}{$
    = \frac{\left\lVert \boldsymbol{W}^{(S)}_{\mathrm{dec},k} \right\rVert_{1}}
           {\left\lVert \boldsymbol{W}^{(O)}_{\mathrm{dec},k} \right\rVert_{1}+\left\lVert \boldsymbol{W}^{(S)}_{\mathrm{dec},k} \right\rVert_{1}+\left\lVert \boldsymbol{W}^{(R)}_{\mathrm{dec},k} \right\rVert_{1}}
$},
\end{equation}
\begin{equation}
    \mathrm{MAS}_{R}
\resizebox{0.73\linewidth}{!}{$
    = \frac{\left\lVert \boldsymbol{W}^{(R)}_{\mathrm{dec},k} \right\rVert_{1}}
           {\left\lVert \boldsymbol{W}^{(O)}_{\mathrm{dec},k} \right\rVert_{1}+\left\lVert \boldsymbol{W}^{(S)}_{\mathrm{dec},k} \right\rVert_{1}+\left\lVert \boldsymbol{W}^{(R)}_{\mathrm{dec},k} \right\rVert_{1}}
$}.
\end{equation}
Here, $O$, $S$, and $R$ denote the original base model, the SFT-tuned model, and the RL-tuned model, respectively.
By construction, $\mathrm{MAS}_{O}$, $\mathrm{MAS}_{S}$, $\mathrm{MAS}_{R}\in[0,1]$, $\mathrm{MAS}_{O} +\mathrm{MAS}_{S} + \mathrm{MAS}_{R}=1$.
The model with the largest MAS value is the one to which the feature is most strongly attributed. For example, a feature with $\mathrm{MAS}_{S}\to1$ is an SFT-specific feature.

\begin{table*}[t]
  \centering
  \vspace{-1mm}
  \setlength{\tabcolsep}{9pt} 
  \resizebox{0.999\textwidth}{!}{
    \begin{tabular}{l|ccc|ccccc}
    \toprule
    \multirow{2}{*}{\textbf{Model}} 
      & \multicolumn{3}{c|}{\textbf{Math Reasoning Tasks}} 
      & \multicolumn{5}{c}{\textbf{General Tasks}} \\
    \cmidrule(lr){2-4} \cmidrule(lr){5-9}
      & \textbf{MATH500} & \textbf{AIME24} & \textbf{AIME25}
      & \textbf{OpenBookQA} & \textbf{CommonsenseQA} & \textbf{HeadQA} & \textbf{SciQ} & \textbf{ARC-Challenge} \\
    \midrule
    Qwen3-4B-Base 
      & 26.0 & 13.3 & 0.0
      & 23.6 & 20.1 & 31.7 & 78.5 & 36.0 \\
    Qwen3-4B-SFT  
      & 68.4 & 13.3 & 13.3
      & 25.8 & 19.6 & 31.2 & 51.8 & 34.4 \\
    Qwen3-4B-RL   
      & 77.0 & 26.7 & 20.0
      & 27.2 & 50.5 & 32.7 & 89.5 & 39.3 \\
    \rowcolor{lightred}
    $\Delta(\text{RL {-} SFT})$ 
      & 8.6 & 13.4 & 6.7
      & 1.4 & 31.0 & 1.5 & 37.7 & 4.9 \\
    \midrule
    Qwen2.5-7B 
      & 40.0 & 10.0 & 3.3
      & 28.4 & 77.6 & 33.7 & 86.9 & 41.9 \\
    Qwen2.5-7B-SFT 
      & 69.2 & 13.3 & 10.0
      & 26.4 & 30.1 & 31.2 & 79.4 & 37.0 \\
    Qwen2.5-7B-RL  
      & 71.4 & 20.0 & 13.3
      & 32.8 & 76.1 & 36.1 & 90.7 & 42.5 \\
    \rowcolor{lightred}
    $\Delta(\text{RL {-} SFT})$ 
      & 2.2 & 6.7 & 3.3
      & 6.4 & 46.0 & 5.0 & 11.3 & 5.5 \\
    \bottomrule
    \end{tabular}
  }
  \caption{Performance comparison of SFT- and RL-tuned models on math reasoning tasks and other tasks.}
  \label{tab:eval_combined}
\end{table*}

\subsection{Identifying Generalization-Controlling Features}
\label{sec:feature_identification}

The final component of our interpretability framework aims to move beyond descriptive comparison and to identify internal features that causally control generalization behavior. Therefore, we hypothesize that RL selectively strengthens a subset of features that play a causal role in enabling generalization across tasks.

Prior work has attempted to localize functionally meaningful features by analyzing which features are frequently activated on specific lexical cues, such as identifying self-reflection features via activations on tokens like ``Wait'', or contrastive features via ``But'' and ``However'' \citep{baek2025towards, galichin2025have}. While such approaches are useful for interpretability, they implicitly assume that functional features are tightly coupled to surface-level tokens. In contrast, we think that \textbf{features controlling generalization should not depend on specific words or task-specific lexical patterns}. Instead, such features should be identifiable through their functional role, namely, whether they systematically alter the model’s decision-making behavior across tasks in situations where generalization actually occurs.

We propose to localize generalization-related features by focusing on samples that explicitly instantiate generalization behavior. For each task, we construct a subset of samples on which the base model fails but the RL-tuned model succeeds. These samples represent the minimal evidence of generalization and are the only instances where generalization can be unambiguously observed. We refer to them as \textbf{generalization-critical samples}.

For each generalization-critical sample, we extract the residual-stream activation at the final token position from the base model and the RL-tuned model, respectively. Then we encode them separately using the model-specific branches of the trained Sparse Crosscoder encoder, yielding feature vectors $\boldsymbol{f}^{(RL)}{(x)}$ and $\boldsymbol{f}^{(Base)}{(x)}$, respectively.

For a feature $k$, we define its \textbf{generalization score} on a given task as the average activation difference between the RL model and the base model across all generalization-critical samples:
\begin{equation}
    \mathrm{Score}_k
    = \mathbb{E}_{x \in \mathcal{G}}
    \left[
    \boldsymbol{f}_k^{(\mathrm{RL})}(x)
    -
    \boldsymbol{f}_k^{(\mathrm{Base})}(x)
    \right],
\end{equation}
where $\mathcal{G}$ denotes the set of generalization-critical samples for that task, and $\boldsymbol{f}_k^{(\cdot)}(x)$ is the activation of feature $k$ for input $x$.
For each task, we retain features whose scores exceed a threshold $t$. These features are treated as task-relevant generalization features. Finally, we take the intersection of the selected feature sets across all tasks, yielding a set of features that consistently contribute to successful generalization regardless of task domain.

In summary, the features we identify are those that (i) systematically differentiate the RL model from the base model on generalization-critical samples, (ii) influence model behavior in a task-agnostic manner, and (iii) do not rely on explicit task knowledge or lexical triggers.

\section{Phenomena: Performance Discrepancies of Reasoning Models}
\label{sec:training}

To ensure that differences between the resulting reasoning models can be attributed solely to the training paradigm rather than confounding factors such as data composition, we deliberately avoid using existing off-the-shelf distilled or RL-tuned models. Instead, we trained both the SFT and RL models from the same base model on an identical dataset using full-parameter tuning, ensuring that any observed differences can be attributed solely to the training paradigms. We performed this controlled tuning on the Qwen-3-4B-Base \cite{yang2025qwen3} and Qwen2.5-7B \cite{yang2025qwen25} models, respectively. Details about training datasets, implementation specifics, and hyperparameters are provided in Appendix~\ref{app:sft_rl_training}.

\paragraph{Evaluation Benchmarks.}
\label{sec:benchmarks}

We evaluated the SFT- and RL-tuned models on a diverse suite of benchmarks, spanning both mathematical reasoning and other general tasks. Specifically, the mathematical reasoning benchmarks include MATH500 \citep{hendrycksmeasuring}, AIME24, and AIME25, while other tasks comprise OpenBookQA \citep{mihaylov2018can}, CommonsenseQA \citep{talmor-etal-2019-commonsenseqa}, HeadQA \citep{vilares-gomez-rodriguez-2019-head}, SciQ \citep{welbl2017crowdsourcing}, and ARC-Challenge \citep{DBLP:journals/corr/abs-1803-05457}. Detailed descriptions of benchmarks and evaluation metrics can be found in Appendix \ref{app:benchmarks}.

\paragraph{Results.} As shown in Table \ref{tab:eval_combined}, we observe that RL-tuned models exhibit strong generalization across diverse domains, whereas SFT-tuned models sometimes suffer from general capabilities forgetting, consistent with observations reported in prior work \cite{huan2025does}.

\begin{figure*}[t]
    \centering
    \includegraphics[width=0.246\textwidth]{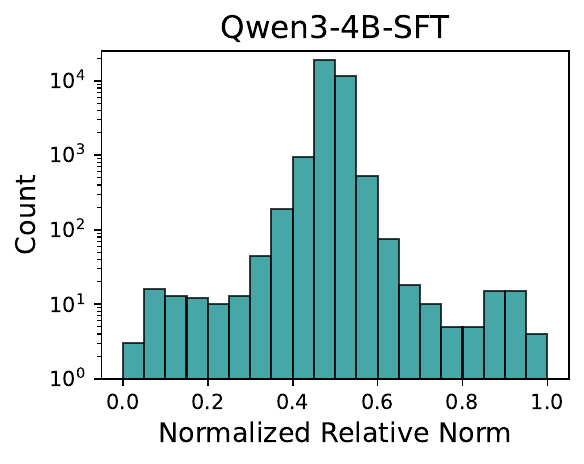}
    \hspace{-2.5mm}
    \includegraphics[width=0.246\textwidth]{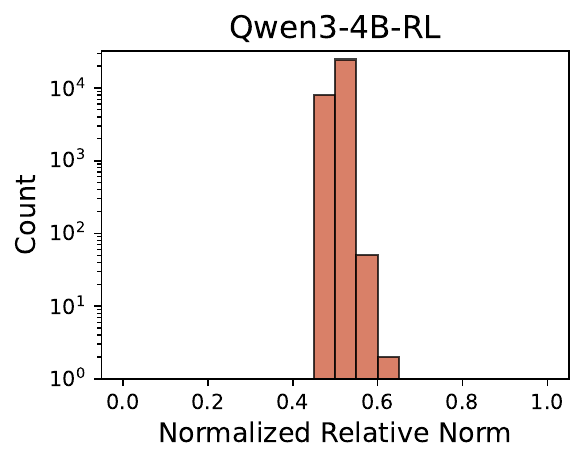}
    \hfill
    \includegraphics[width=0.246\textwidth]{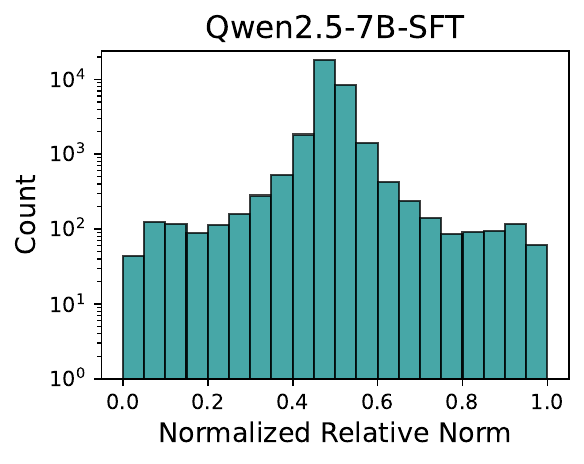}
    \hspace{-2.5mm}
    \includegraphics[width=0.246\textwidth]{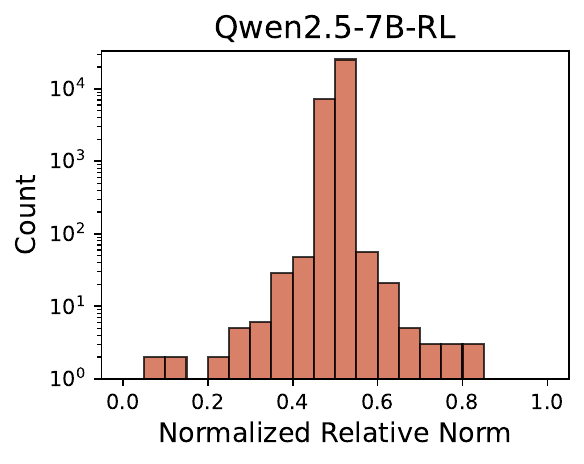}
    \caption{Distribution of Normalized Relative Norms across different training methods and different model scales.}
    \label{fig:NRN}
\end{figure*}

\section{Experiments: Comparing SFT and RL Models via Two-Model Sparse Crosscoders} 

To analyze how SFT and RL alter the internal representations of the base model, we independently trained two Sparse Crosscoders: one between the SFT-tuned model and the base model, and another between the RL-tuned model and the base model. Implementation and training details are provided in Appendix~\ref{app:crossder_training}.

\subsection{Feature Discrepancies Between RL and SFT}

Using the trained crosscoders, we computed the NRNs for both the SFT- and the RL-tuned models relative to the original base model, denoted as $\mathrm{NRN}_{\mathrm{SFT}}$ and $\mathrm{NRN}_{\mathrm{RL}}$, respectively. The resulting NRN distributions are visualized in Figure~\ref{fig:NRN}.

Across both training paradigms, the majority of features cluster around NRN $\approx 0.5$, indicating that most features are shared between the fine-tuned models and the base model. Moreover, the number of features decays approximately exponentially toward both extremes of the distribution, suggesting that highly model-specific features are relatively rare. However, the tails of the distribution, corresponding to model-specific features, exhibit strikingly different behaviors for SFT and RL.

\paragraph{SFT Induces a Large Number of Unique Features.} 
For the SFT-tuned model, the NRN distribution exhibits a pronounced right tail, with a substantial number of features achieving $\mathrm{NRN}_{\mathrm{SFT}}>0.8$, and some even approaching $\mathrm{NRN}_{\mathrm{SFT}}=1$. These features correspond to representations that are almost entirely unique to the SFT model.
At the same time, we also observe a non-negligible number of features with $\mathrm{NRN}_{\mathrm{SFT}}\to0$, indicating features that are effectively exclusive to the original model. These two extremes suggest that SFT induces a pronounced representational shift: while introducing many new, highly specialized features, it also suppresses or abandons a considerable portion of the base model’s feature repertoire. 

Furthermore, the Qwen2.5-7B crosscoders exhibit a larger number of features with extreme NRN values than the Qwen3-4B-Base counterparts, implying that larger models tend to develop a richer set of distinctive internal features during post-training.

\paragraph{RL Preserves Core Representations While Introducing Fewer, Milder Deviations.} 

In contrast, the RL-tuned model exhibits a markedly different NRN profile. Only a small number of features exhibit extreme $\mathrm{NRN}_{\mathrm{RL}}$ values (close to either 0 or 1), with very few approaching either end of the spectrum. This indicates that RL introduces relatively few novel, model-specific features and largely preserves the original model’s representational structure. 
This pattern can be attributed to the nature of RL supervision. Unlike SFT, which directly constrains the entire reasoning trajectory, RL only provides outcome-level feedback based on final answer correctness. As a result, RL does not force the model to adopt a particular reasoning style or surface form. Instead, it selectively reinforces internal computations that contribute to correct decisions, while leaving much of the base model’s representational structure intact.

Overall, these results suggest a fundamental contrast in post-training dynamics: SFT drives substantial feature turnover, generating many highly model-specific features while erasing others, whereas RL induces restrained and targeted adjustments that largely preserve the original feature space.

\begin{figure*}[t]
    \centering
    \includegraphics[width=0.246\textwidth]{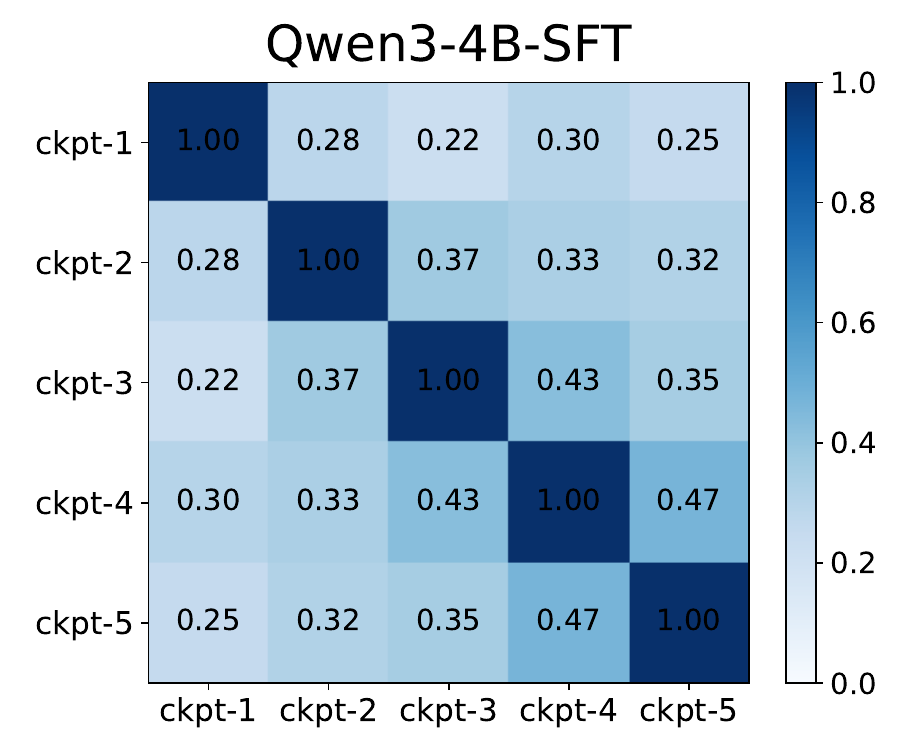}
    \hspace{-2.5mm}
    \includegraphics[width=0.246\textwidth]{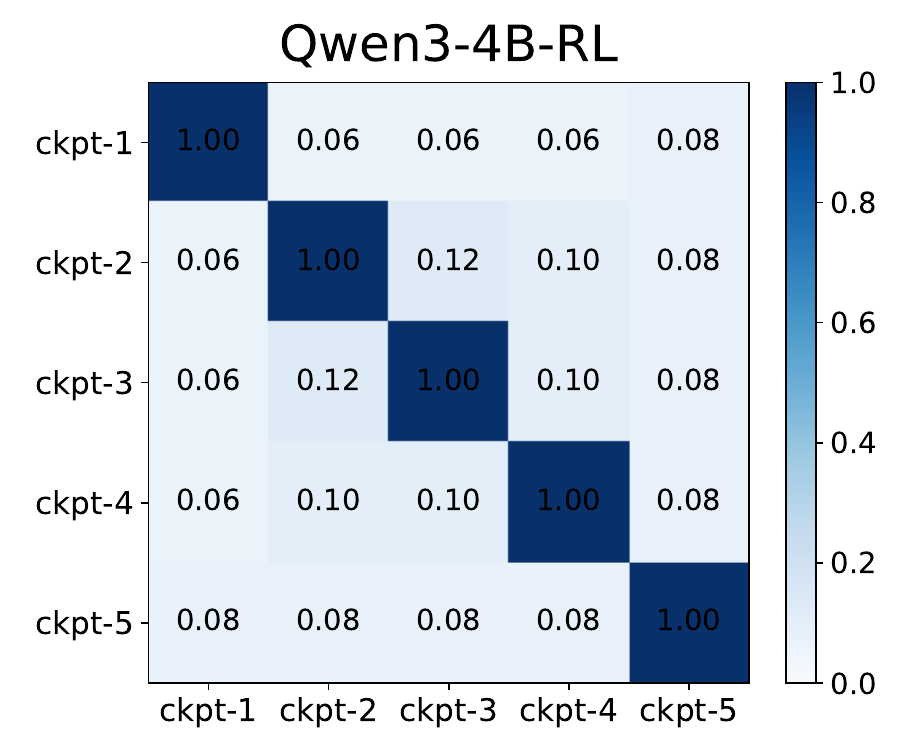}
    \hspace{1mm}
    \includegraphics[width=0.246\textwidth]{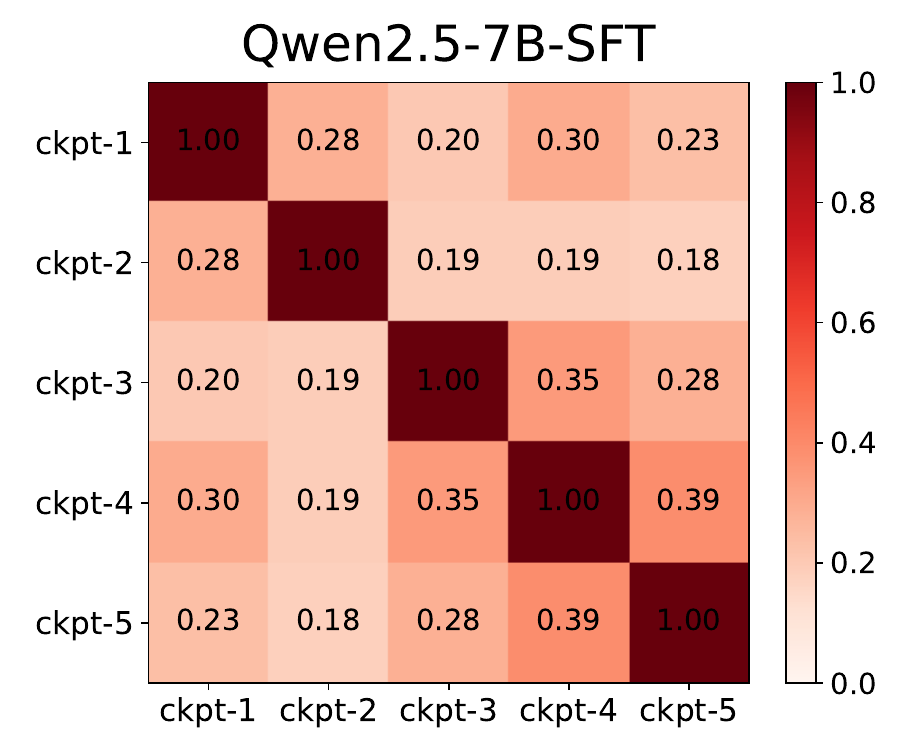}
    \hspace{-2.5mm}
    \includegraphics[width=0.246\textwidth]{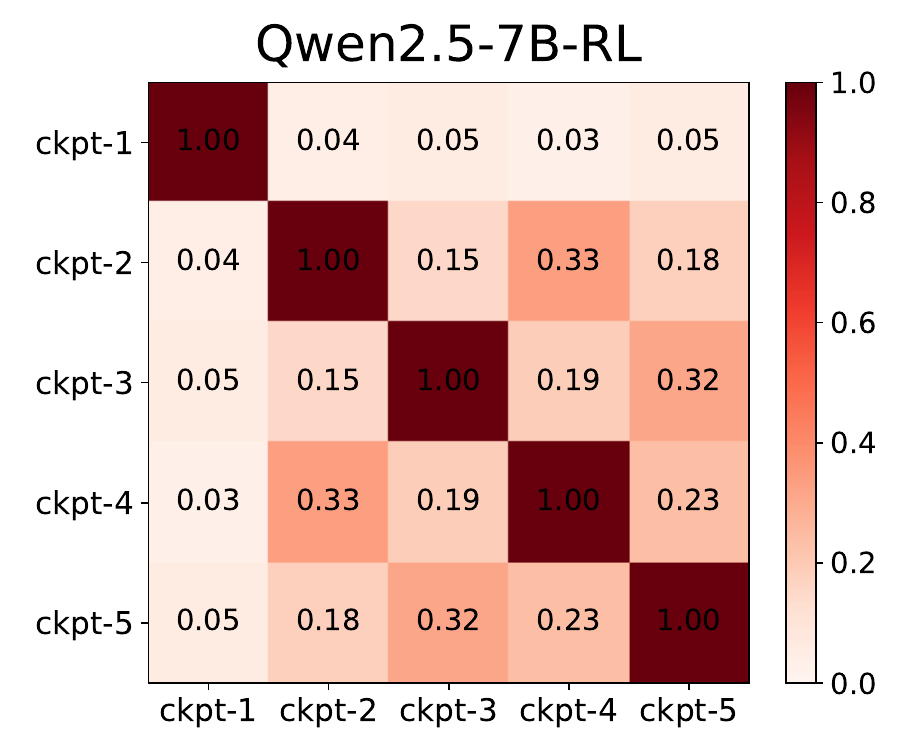}
    \vspace{-1mm}
    \caption{Feature overlap heatmaps across training checkpoints under different training paradigms.}
    \label{fig:overlap}
    \vspace{-1mm}
\end{figure*}

\begin{figure*}[t]
    \centering
    \includegraphics[width=0.88\textwidth]{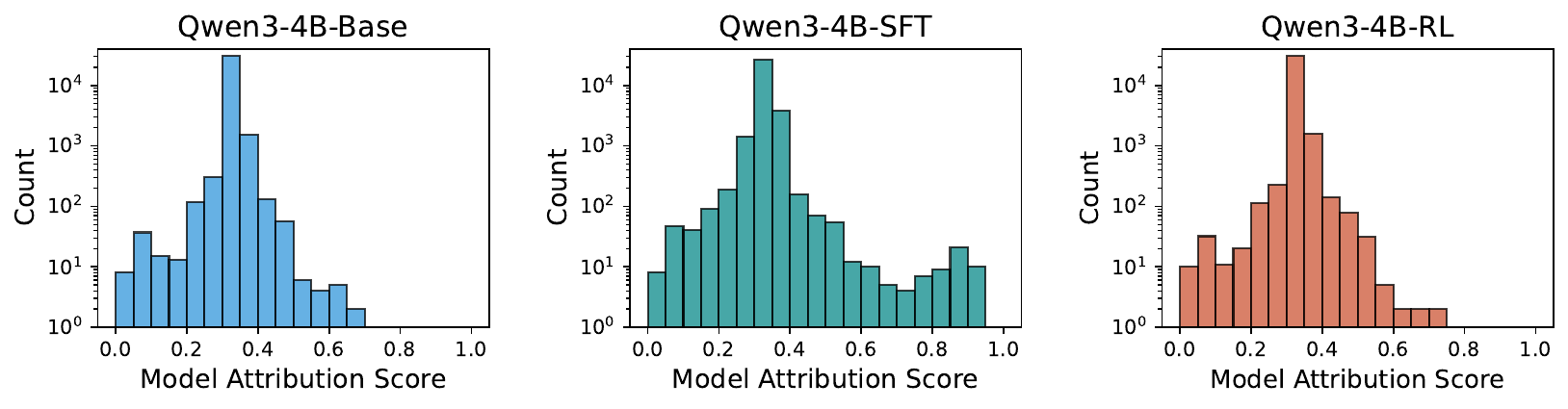}
    \caption{Distribution of Model Attribution Scores across different training methods on Qwen3-4B-Base.}
    \label{fig:MAS_4b}
    \vspace{-2mm}
\end{figure*}

\subsection{Feature Formation and Evolution During Training} 
To further understand how SFT and RL alter the internal representations of the base model over time, we analyzed the temporal evolution of features during training. Rather than focusing solely on the final tuned models, we examined how model-specific features emerge, persist, and change across training checkpoints.

\paragraph{Experimental Setup.}

For each training paradigm, we saved checkpoints at regular intervals (every one-fifth of an epoch) and pair each checkpoint with the base model to train a Sparse Crosscoder, yielding five crosscoders per paradigm. This setup enables stage-wise comparisons between the base model and partially trained models. Using these crosscoders, we conducted two complementary analyses: (1) \textbf{feature overlap across checkpoints}, and (2) \textbf{feature rank shifts between consecutive checkpoints}. These analyses allow us to quantify both the stability of learned features and the extent to which the internal feature space is reorganized during training.

\paragraph{SFT Quickly Establishes a Relatively Fixed Set of Internal Features, While RL Promotes a Slower and More Incremental Process of Feature Formation.}
For each checkpoint, we ranked features by descending NRN and retain the top 50 as the most distinctive (i.e., model-specific) features at that stage. We then computed the overlap of these top-ranked features across different checkpoints. As shown in Figure~\ref{fig:overlap}, the SFT-tuned model exhibits a high degree of feature overlap across checkpoints. A substantial portion of the top-ranked SFT-specific features identified at early checkpoints persists throughout later stages of training. This indicates that many of the features introduced by SFT emerge early and remain consistently dominant during subsequent optimization. 

In contrast, the RL-tuned model displays negligible overlap between the top-ranked feature sets of adjacent checkpoints. Features that are highly ranked at an earlier checkpoint rarely remain among the top features at the subsequent checkpoint. Only toward the end of training does the overlap increase, suggesting that RL delays feature consolidation and instead explores a broader set of candidate features before gradually stabilizing its feature composition.

\paragraph{SFT Primarily Refines Feature Strengths After Early Formation, While RL Continuously Reorders Feature Importance}

Beyond feature persistence, we further examined how the relative importance of features evolves by analyzing rank shifts between consecutive checkpoints. We observe that SFT rapidly forms a comparatively stable hierarchy of model-specific features, with later training primarily refining their relative importance. In contrast, RL induces features more gradually, with training persistently reshaping which internal features are most salient for generating correct outcomes. Visualization results and detailed analyses are provided in Appendix~\ref{app:rank_shift}.

To assess whether our findings generalize across model families, we further conduct controlled post-training and feature-level analysis on a different architecture, namely Llama3.1-8B-Instruct \cite{grattafiori2024llama}. The corresponding results are presented in Appendix~\ref{sec:llama3}. We observe consistent patterns with those found in the Qwen family, indicating that the representational distinctions between SFT and RL are not model-specific but broadly applicable across architectures.

In conclusion, the contrast between SFT and RL features reveals two fundamentally different modes of representation change:

\begin{figure*}[t]
    \centering
    \includegraphics[width=0.41\textwidth]{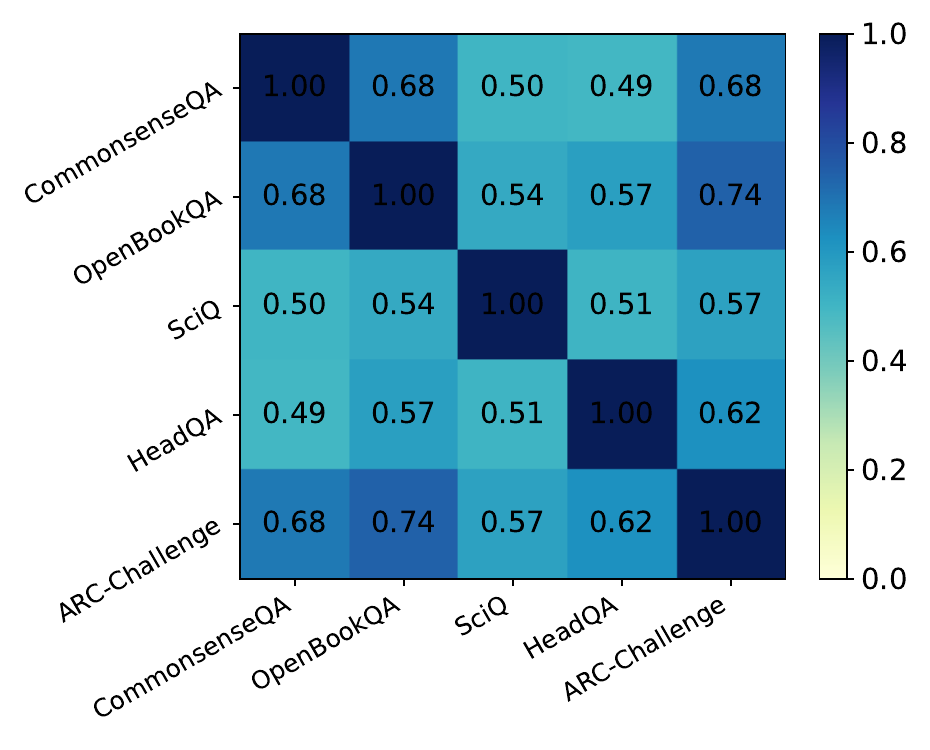}
    \includegraphics[width=0.41\textwidth]{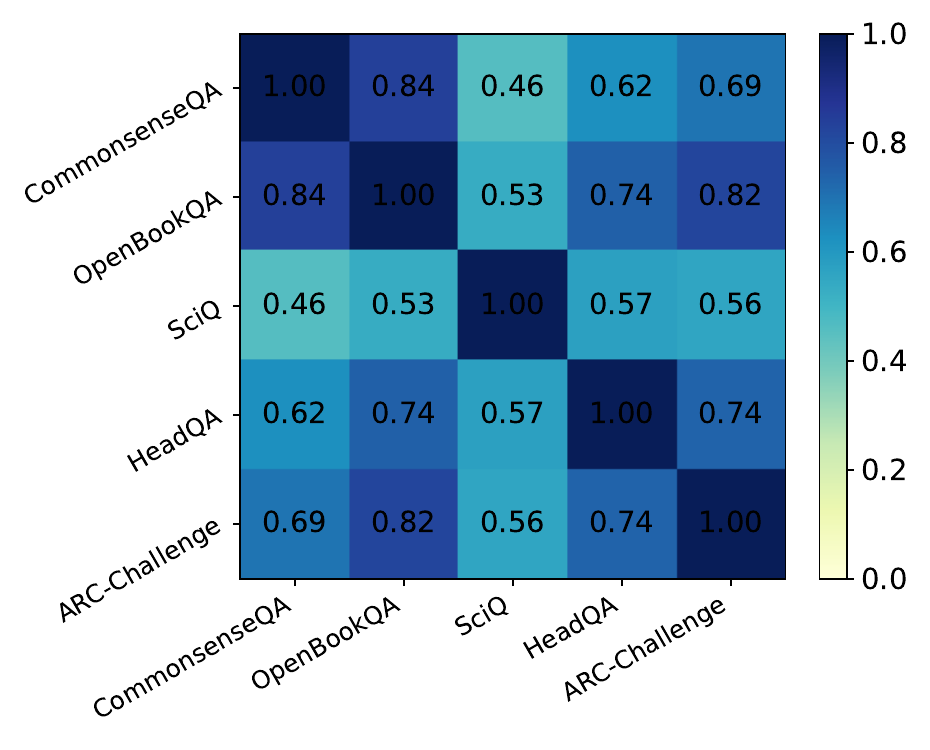}
    \vspace{-2mm}
    \caption{Overlap of identified generalization-controlling features across tasks. Left: Qwen3-4B-RL vs.\ Qwen3-4B-Base. Right: Qwen2.5-7B-RL vs.\ Qwen2.5-7B.}
    \label{fig:gena_feature_overlap}
\end{figure*}

\begin{itemize}
    \item
    SFT rapidly introduces a large number of model-specific features that closely replicate the teacher’s reasoning traces, while simultaneously discarding many pre-existing features. After the formation, the overall feature composition remains largely stable. Subsequent optimization primarily adjusts the relative strengths of these features. This yields a rigid and specialized feature space tightly aligned with the training distribution.
    
    \item
    RL introduces feature deviations in a gradual and restrained manner, yielding relatively few and less exclusive features. Rather than substantially replacing the base model’s internal representational structure, RL induces limited feature innovation while exhibiting sustained feature turnover throughout training.
\end{itemize}

These findings indicate that SFT and RL differ not only in the final features they produce, but also in the temporal dynamics of feature formation. Importantly, the prolonged feature reconfiguration observed in RL suggests a training process that remains sensitive to performance feedback throughout optimization, a property that may be closely related to the stronger cross-task generalization behavior analyzed in the next section.

\begin{table*}[t]
  \centering
  \setlength{\tabcolsep}{12pt}
  \resizebox{0.9\textwidth}{!}{
    \begin{tabular}{lccccc}
    \toprule
    \textbf{Model} & \textbf{OpenBookQA} & \textbf{CommonsenseQA} & \textbf{HeadQA} & \textbf{SciQ} & \textbf{ARC-Challenge} \\
    \midrule
    Qwen3-4B-RL & -46.2  & -43.9  & -21.2  & -14.0  & -33.3  \\
    Qwen2.5-7B-RL & -21.9  & -24.4  & -23.8  & -44.4  & -20.0  \\
    \bottomrule
    \end{tabular}%
  }
  \caption{Performance degradation induced by the removal of generalization-related features in RL–tuned models.}
  \label{tab:zero}%
\end{table*}%

\begin{table*}[t]
  \centering
  \setlength{\tabcolsep}{12pt}
  \resizebox{0.9\textwidth}{!}{
    \begin{tabular}{lccccc}
    \toprule
    \textbf{Model} & \textbf{OpenBookQA} & \textbf{CommonsenseQA} & \textbf{HeadQA} & \textbf{SciQ} & \textbf{ARC-Challenge} \\
    \midrule
    Qwen3-4B-Base & +36.3  & +36.0  & +21.2  & +38.0  & +33.3  \\
    Qwen2.5-7B & +12.5  & +24.4  & +14.3  & +55.6  & +40.0  \\
    \bottomrule
    \end{tabular}%
  }
  \caption{Performance improvement achieved by amplifying generalization-related features in the base model.}
  \label{tab:amplify}%
  \vspace{-1mm}
\end{table*}%

\section{Experiments: Joint Comparison with Three-Model Sparse Crosscoders}

Using the trained three-model Sparse Crosscoder, we computed the MASs for each feature, measuring how strongly the feature is attributed to the base, SFT-tuned, or RL-tuned model within a shared feature space. Figure~\ref{fig:MAS_4b} illustrates the MAS distributions for Qwen3-4B-Base across the different training conditions. 
The MAS distributions for the SFT-tuned models exhibit a pronounced right tail. A substantial number of features attain high attribution scores. This pattern suggests that SFT introduces a large set of strongly model-specific features. Compared to SFT, RL produces far fewer features that are strongly attributed to itself. 
The same observations can be found on Qwen2.5-7B, as shown in Figure~\ref{fig:MAS_7b} in Appendix~\ref{app:mas_7b}.

The three-model Sparse Crosscoder provides direct empirical evidence that SFT and RL reshape the internal feature landscape in fundamentally different ways. SFT introduces a large number of strongly model-specific features, while RL predominantly preserves and reweights shared features within the existing representational space. More importantly, it enables downstream mechanistic analysis: it serves as the foundation for identifying features associated with cross-task generalization in the next section.

\section{Mechanistic Explanation for the Generalization Ability of RL}

RL-tuned models are optimized solely on mathematical reasoning tasks, yet they achieve substantial gains not only on in-domain mathematics benchmarks but also on disparate tasks such as commonsense and scientific knowledge question answering.
This cross-task generalization cannot be explained by data overlap or task similarity, and therefore calls for a mechanistic explanation. 

Our prior analyses provide an initial explanation: RL induces restrained, incremental feature changes, whereas SFT introduces aggressive feature reconfiguration that may lead to over-specialization. 
However, this explanation remains indirect. To establish a more concrete mechanistic account, we seek direct evidence linking specific internal features to the observed generalization behavior.

\paragraph{Experimental Setup.}
We performed feature identification using the method proposed in Section \ref{sec:feature_identification} on all five evaluation tasks introduced in Section \ref{sec:benchmarks}. For each task, we set the threshold $t$ as 20\% of the maximum score observed for that task. When examining the generalization features identified for each task, we observe a striking degree of consistency across tasks. To further substantiate this observation, we quantified the overlap among the identified features across different tasks, as illustrated in Figure~\ref{fig:gena_feature_overlap}. The results reveal a substantial level of feature overlap, even over 80\%. Despite the diversity of these benchmarks, many features consistently appear across tasks, strongly suggesting the existence of task-agnostic features that support generalization.

The final intersection across all tasks contains 50 features for Qwen3-4B-Base and 16 features for Qwen2.5-7B, which we refer to as the final generalization-controlling features.

\subsection{Causal Validation via Feature Interventions}

To verify the functional importance of these features, we conducted two complementary intervention experiments.

\paragraph{Zeroing Generalization Features in the RL Models.}

We first conducted ablation experiments by zeroing out the identified features in the RL model and evaluating performance on the corresponding generalization-critical samples. As shown in Table~\ref{tab:zero}, this intervention leads to substantial performance degradation. In particular, for OpenBookQA and CommonsenseQA, zeroing these features causes the Qwen3-4B-RL model to answer over 40\% of previously correct samples incorrectly. These results indicate that the identified features are necessary for successful generalization. 

\paragraph{Amplifying Generalization Features in the Base Models.} Conversely, we amplified the same set of features in the base model by setting their activations to a fixed large value (we set to 3.0), following standard feature intervention practices in SAE-based analyses \cite{zhang2025large, han-etal-2025-towards}. The results are shown in Table~\ref{tab:amplify}.
Across all tasks, performance improves substantially. In particular, for SciQ, amplifying these features enables Qwen2.5-7B to correctly answer 56\% of samples that it previously answered incorrectly.

This finding suggests that the base model does not lack the necessary knowledge. Instead, the relevant upstream circuitry is present but not naturally activated. Once a control signal is forcibly injected, generalized behaviors emerge prominently.

\paragraph{Generalization to Unseen Tasks.}
We further evaluated whether the identified generalization features transfer beyond the tasks used for feature identification. Specifically, we tested them on two unseen benchmarks, LogiQA \cite{liu2021logiqa} and PIQA \cite{bisk2020piqa}. We observe consistent performance degradation when zeroing these features in RL-trained models, and corresponding performance improvements when amplifying them in base models. This provides additional evidence that the identified features capture a general-purpose generalization mechanism, rather than task-specific heuristics. Detailed experimental setup and results are provided in Appendix~\ref{app:unseen_tasks}.

\section{Conclusion}

This work has investigated why RL-tuned LLMs generalize beyond their training distribution, while SFT often leads to the loss of general capabilities. Using a controlled experimental setup and a feature-level interpretability framework, we have compared how RL and SFT reshape internal representations during post-training. We have shown that SFT rapidly introduces highly specialized features that stabilize early in training, whereas RL induces more restrained and continually evolving feature changes that largely preserve the base model’s representations. Building on this distinction, we have identified a compact set of internal features that causally control cross-task generalization. Feature-level interventions have confirmed their role: disabling these features degrades RL performance, while amplifying them transfers generalization behavior to base models, including on unseen tasks. Overall, our results have provided a mechanistic explanation for why RL generalizes while SFT memorizes, and have demonstrated the value of feature-level interpretability for understanding post-training dynamics in LLMs.

\newpage
\clearpage

\section*{Limitations}

First, our feature-level analysis relies on Sparse Crosscoders as the underlying alignment mechanism. Although this approach enables interpretable and comparable feature representations, the learned feature space is not guaranteed to capture all functionally relevant internal structures. Other interpretability methods or alignment schemes may reveal complementary mechanisms that are not captured by sparse feature decomposition. Second, 
while our inference-time feature interventions establish causal links to model behavior, they do not directly inform how to design training objectives that explicitly encourage generalization-controlling features. We leave the development of such training strategies to future work.

\section*{Acknowledgements}
The present research was supported by the National Key Research and Development Program of China (Grant No. 2024YFE0203000) and the International Cooperation Program for Innovative Talents Development of CSC (Grant No. CXXM2310203712). We would like to thank the anonymous reviewers for their insightful comments.


\bibliography{custom}

\newpage
\clearpage

\appendix

\section{Experimental Settings}
\label{sec:detailed_Exp_setting}
In this section, we provide more details about the experimental settings.

\subsection{Training Details of Crosscoders}
\label{app:crossder_training}

Following \citet{baek2025towards}, we train each crosscoder with $d_{sparse}=32,768$ features on 200 million tokens from open-thoughts/OpenThoughts-114k\footnote{https://huggingface.co/datasets/open-thoughts/OpenThoughts-114k} and another 200 million tokens from togethercomputer/RedPajama-Data-1T-Sample\footnote{https://ai.gitee.com/hf-datasets/togethercomputer/RedPajama-Data-1T-Sample} dataset. The former includes math, science, and code reasoning traces generated by DeepSeek-R1, whereas the latter contains general-domain texts. To avoid distributional bias, the two datasets are merged and jointly shuffled before training, rather than being trained sequentially. This mixed corpus allows the crosscoder to capture both reasoning-related and general linguistic features. All crosscoders are trained to reconstruct the residual stream of the middle layer of each model. Training is conducted with a batch size of 1024 and a learning rate of $1 \times 10^{-4}$. The $\beta$, which controls the sparsity regularization strength, is set to 2.

\subsection{Training Setup for SFT and RL}
\label{app:sft_rl_training}
This subsection details the training setup used for SFT and RL, including the datasets, optimization objectives, implementation details, and training hyperparameters for both paradigms.

\paragraph{Training Datasets}

To train our models, we adopt the high-quality mathematics dataset constructed by \citet{huan2025does}, which consists of 47K high-quality mathematics problems derived from MATH \citep{hendrycksmeasuring} and DeepScaler \citep{deepscaler2025}. Both training paradigms are applied to the same backbone, Qwen3-4B-Base and Qwen2.5-7B. For RL, the model is optimized using standard Group Relative Policy Optimization (GRPO, \citealp{shao2024deepseekmath}), where rewards are computed by comparing the model’s final answers against the gold answers provided in the dataset. This setup strictly supervises outcome correctness without exposing intermediate reasoning traces to the model. For SFT, the training targets are complete chain-of-thought (CoT) reasoning traces generated by a strong teacher model, Qwen3-32B-Instruct \citep{yang2025qwen3}, and we retain only those responses that lead to correct final answers.

\subsection{Training Details}

\paragraph{Reinforcement Learning} (RL) has recently demonstrated notable success in enhancing the complex, multi-step reasoning capabilities of LLMs by optimizing policies with scalar reward signals. In our study, we adopt the \texttt{verl} framework \citep{sheng2025hybridflow} and implement GRPO \citep{shao2024deepseekmath} on the Qwen-3-4B-Base and Qwen2.5-7B models. Training is performed with an overall batch size of 128 and a learning rate of $1 \times 10^{-6}$. We set the generation sequence length up to 16k tokens, and perform 8 rollouts per prompt, updating the model in mini-batches of 64 samples. Clipping thresholds are set between 0.22 and 0.28 to ensure stable policy updates, while both KL-divergence and entropy penalties are turned off (coefficients set to zero). The model is trained for one epoch, and the checkpoint from the final iteration is preserved for subsequent evaluation. 

\paragraph{Supervised Fine-Tuning} (SFT) remains a widely adopted approach for transferring knowledge and desired behaviors from large pre-trained language models to task-adapted or resource-constrained models. In particular, SFT has recently been extensively used as a form of reasoning distillation, where the intermediate reasoning processes of a stronger teacher model are distilled into a smaller student model through high-quality, teacher-generated chain-of-thought annotations \citep{guo2025deepseek}. By minimizing the cross-entropy loss on curated datasets, SFT allows the model to internalize desired behaviors and reasoning patterns in a fully supervised manner. In our experiments, we employ the \texttt{LLaMA-Factory} framework \citep{zheng2024llamafactory} to fine-tune the two models on teacher-provided chain-of-thought traces. The learning rate is set to $5 \times 10^{-5}$ with a batch size of 128. For consistency with the RL setting, training is also performed for one epoch, and the final checkpoint is retained for downstream evaluation. 

All training runs were conducted on H200 and H100 GPUs.

\subsection{Detailed Description of the Benchmarks and Evaluation Metrics}
\label{app:benchmarks}

In the experiment, we evaluated our models on a broad range of benchmarks designed to probe different aspects of reasoning and generalization. These benchmarks can be broadly grouped into mathematical reasoning tasks and non-mathematical tasks, which vary in domain knowledge and reasoning requirements.

\paragraph{Math reasoning tasks}
The following benchmarks primarily evaluate a model’s ability to perform explicit multi-step mathematical reasoning, often requiring symbolic manipulation and structured problem solving.

\begin{itemize}
    \item \textbf{MATH500} \citep{hendrycksmeasuring}: A subset of 500 problems sampled from the MATH dataset, covering algebra, geometry, number theory, and combinatorics. Each problem typically requires multi-step derivations and precise numerical or symbolic answers.

    \item \textbf{AIME24 / AIME25}: Problems drawn from the 2024 and 2025 editions of the American Invitational Mathematics Examination (AIME). Each benchmark consists of 30 challenging short-answer questions that demand careful reasoning and mathematical insight.

\end{itemize}

\paragraph{Other tasks}

The following benchmarks assess reasoning and knowledge use outside the mathematical domain, making them particularly suitable for evaluating generalization.

\begin{figure*}[t]
    \centering
    \includegraphics[width=0.92\textwidth]{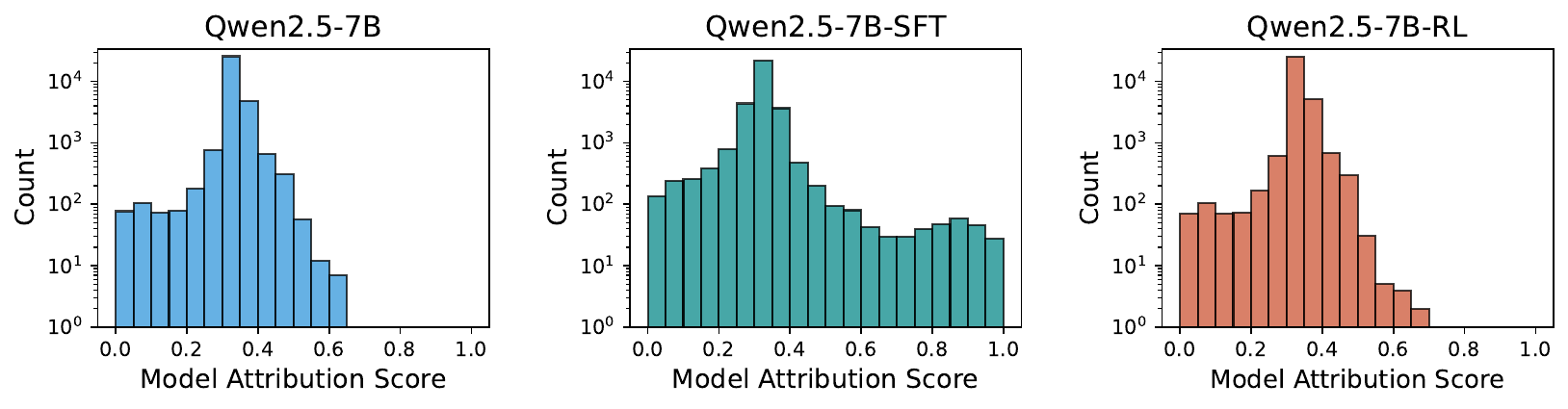}
    \caption{Distribution of Model Attribution Scores across different training methods on Qwen2.5-7B.}
    \label{fig:MAS_7b}
\end{figure*}

\begin{figure}[t]
    \centering
    \includegraphics[width=0.5\linewidth]{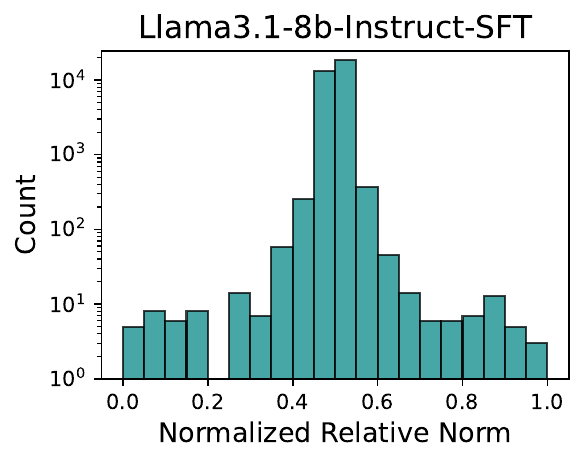}
    \hspace{-2.5mm}
    \includegraphics[width=0.5\linewidth]{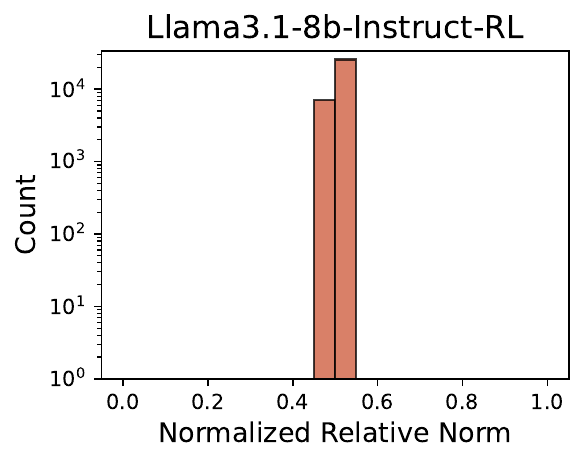}
    \vspace{-1mm}
    \caption{Distribution of Normalized Relative Norms across different training methods on Llama3.1-8B-Instruct.}
    \label{fig:llama_NRN}
\end{figure}

\begin{itemize}

    \item  \textbf{OpenBookQA} \citep{mihaylov2018can}: A multiple-choice question answering benchmark focused on elementary science knowledge. Each question is associated with a set of core facts, and the model must select the correct answer from four options.
    
    \item \textbf{CommonsenseQA} \citep{talmor-etal-2019-commonsenseqa}:  
    A multiple-choice benchmark designed to test general commonsense knowledge. Questions are constructed around concepts from structured knowledge bases, with distractor options chosen to be semantically plausible.
    
    \item \textbf{HeadQA} \citep{vilares-gomez-rodriguez-2019-head}:  
    A medical question answering benchmark composed of multiple-choice questions derived from healthcare specialization exams, including questions across pharmacology, chemistry, nursing, psychology, biology, and medicine.
    
    \item \textbf{SciQ} \citep{welbl2017crowdsourcing}:  
    A science question answering dataset focusing on elementary-level scientific concepts.
    
    \item \textbf{ARC-Challenge} \citep{DBLP:journals/corr/abs-1803-05457}:  
    The more difficult subset of the AI2 Reasoning Challenge, containing grade-school science questions that typically require multi-hop reasoning and background knowledge integration.

\end{itemize}

Collectively, these benchmarks span a wide range of domains and reasoning types. Their diversity allows us to probe whether improvements induced by RL training reflect task-specific adaptations or general-purpose reasoning capabilities that transfer across domains.

We used \texttt{LLM-Evaluation-Harness} \cite{eval-harness} to evaluate the models’ performance on OpenBookQA, CommonsenseQA, HeadQA, SciQ, and ARC-Challenge, and used \texttt{Eval-Chemy} \cite{Evalchemy-Automatic-evals-for-LLMs} to evaluate the performance on MATH500, AIME24, and AIME25. In our experiments, we adopted exact-match accuracy to evaluate the models’ performance on math reasoning tasks. Specifically, for AIME24 and AIME 25, we averaged accuracy on 10 repetitions. For MATH 500, our score is the average accuracy over 3 repetitions. For other benchmarks, we report accuracy following standard evaluation protocols.

\section{More Experimental Results}
In this section, we present additional experimental results that complement the main findings discussed in the main paper.

\subsection{Feature Rank Shifts Between Consecutive Checkpoints}
\label{app:rank_shift}
To further examine the dynamics of feature evolution during training, for each pair of adjacent checkpoints, we compute how much each feature’s rank (based on NRN) changes from one checkpoint to the next.

\begin{figure*}[t]
    \centering
    \includegraphics[width=\textwidth]{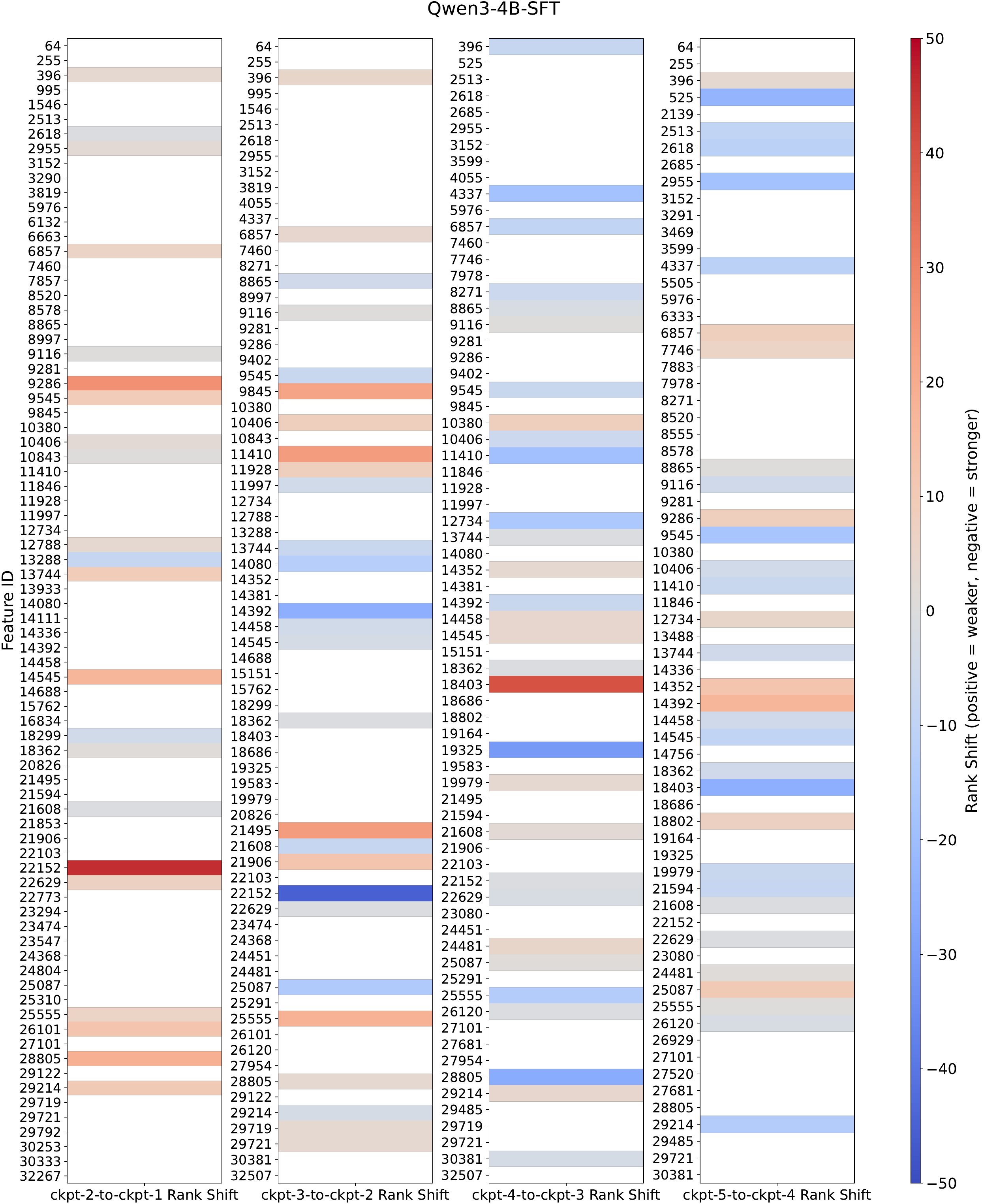}
    \caption{Feature rank shift across adjacent checkpoints during SFT on Qwen3-4B-Base.}
    \label{fig:rank_shift_4b_sft}
\end{figure*}

\begin{figure*}[t]
    \centering
    \includegraphics[width=\textwidth]{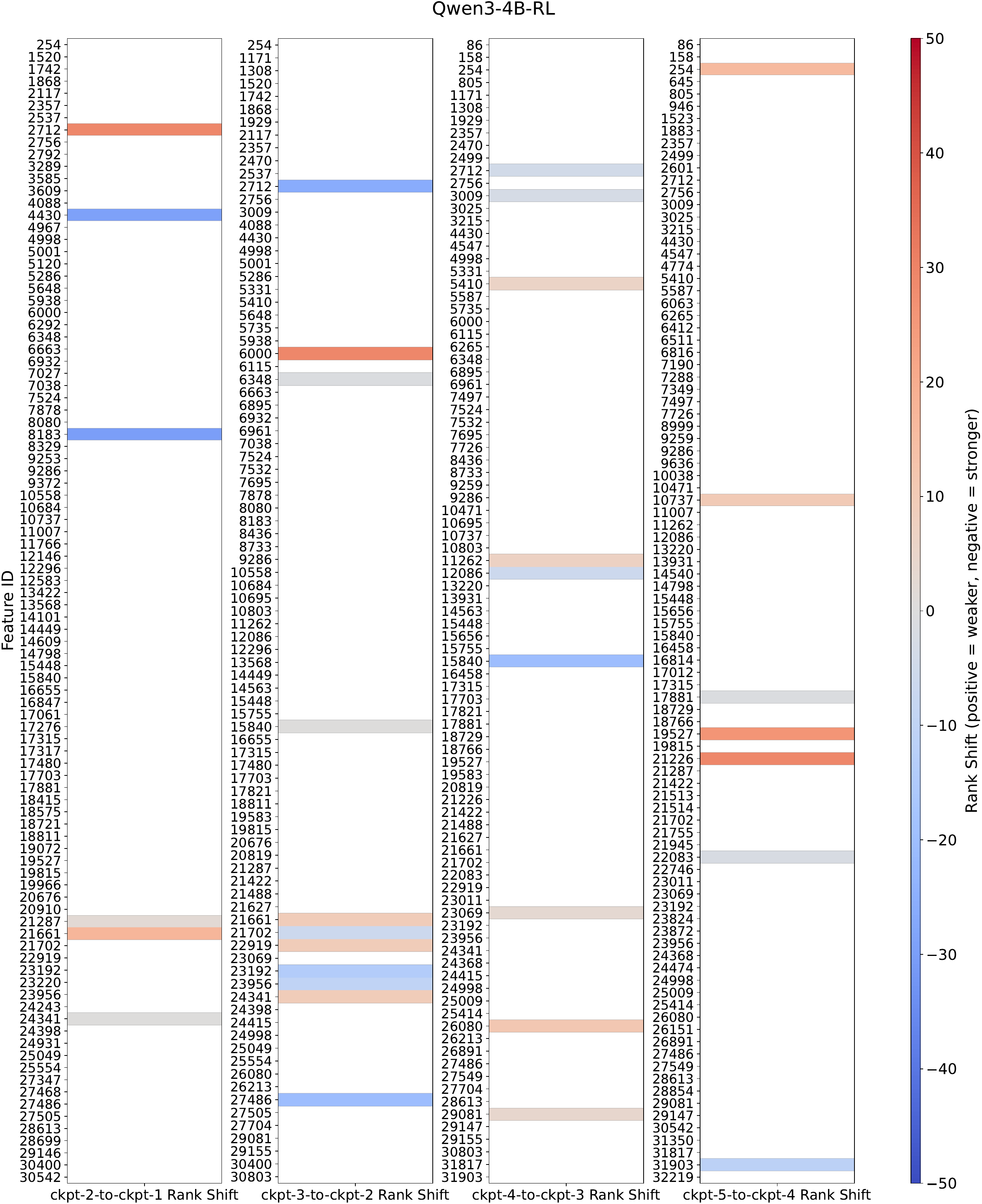}
    \caption{Feature rank shift across adjacent checkpoints during RL on Qwen3-4B-Base.}
    \label{fig:rank_shift_4b_rl}
\end{figure*}

\begin{figure*}[t]
    \centering
    \includegraphics[width=\textwidth]{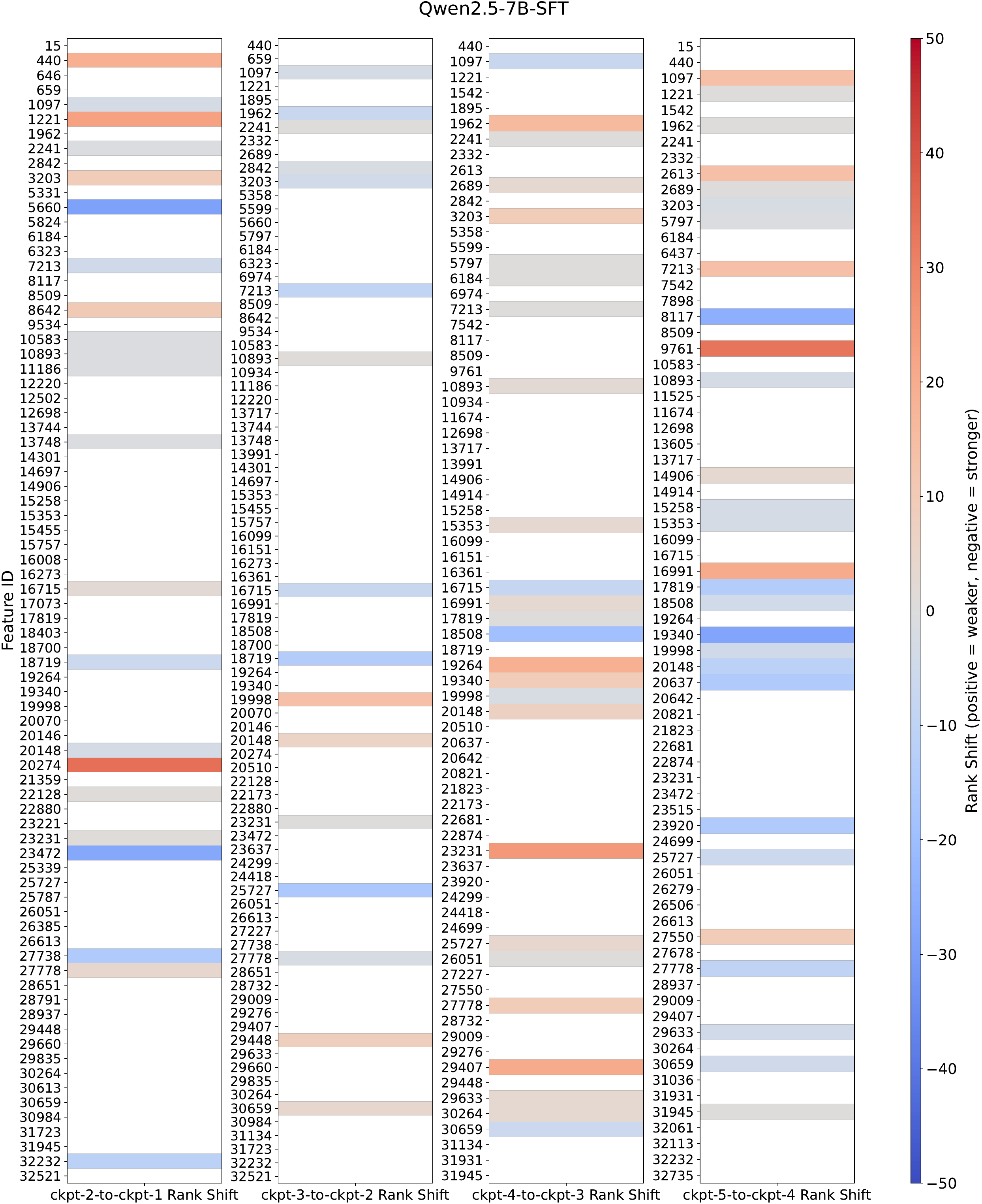}
    \caption{Feature rank shift across adjacent checkpoints during SFT on Qwen2.5-7B.}
    \label{fig:rank_shift_7b_sft}
\end{figure*}

\begin{figure*}[t]
    \centering
    \includegraphics[width=\textwidth]{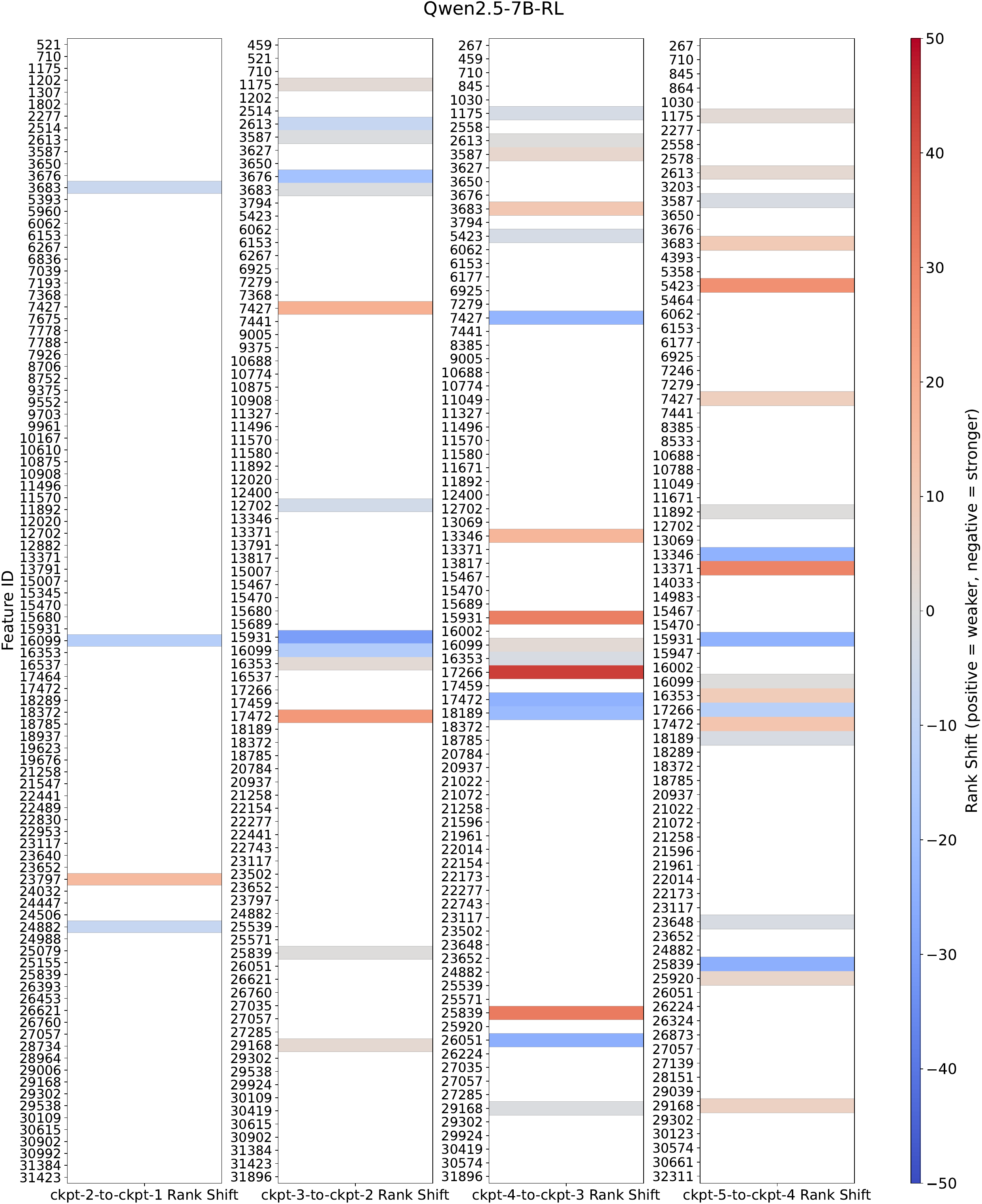}
    \caption{Feature rank shift across adjacent checkpoints during RL on Qwen2.5-7B.}
    \label{fig:rank_shift_7b_rl}
\end{figure*}

The results, visualized in Figure~\ref{fig:rank_shift_4b_sft} to \ref{fig:rank_shift_7b_rl}, reveal distinct patterns for SFT and RL. In the SFT-tuned model, rank changes between consecutive checkpoints are generally small. First, the number of blank features is limited, where a blank feature denotes one that appears in the top 50 at one checkpoint but falls outside the top 50 at the adjacent checkpoint. Second, most features exhibit relatively small rank shifts, as reflected by lighter colors in the visualization. Moreover, some features consistently appear in the top 50 across all checkpoints, indicating that certain SFT-induced features are established early and persist throughout training. Together, these observations suggest that SFT quickly establishes a relatively stable hierarchy of model-specific features, with later training primarily refining their relative importance.

By contrast, the RL-tuned model exhibits substantially larger rank shifts across checkpoints. A large number of features are blank between adjacent checkpoints, indicating frequent turnover among top-ranked features. Features frequently undergo significant reordering, with previously prominent features diminishing in importance and new features rising to prominence at later stages. This pattern suggests that RL-induced features are formed gradually and slowly, with training continuously adjusting which internal features are most relevant for producing correct outcomes.

\subsection{Results on Llama3.1-8B-Instruct}
\label{sec:llama3}
To evaluate the robustness and generality of our findings across model families, we extend our analysis to Llama3.1-8B-Instruct. Following the same controlled experimental setup described in Section \ref{sec:training}, we train both SFT- and RL-tuned models from the same base model using identical data. 

\begin{figure}[t]
    \centering
    \includegraphics[width=0.5\linewidth]{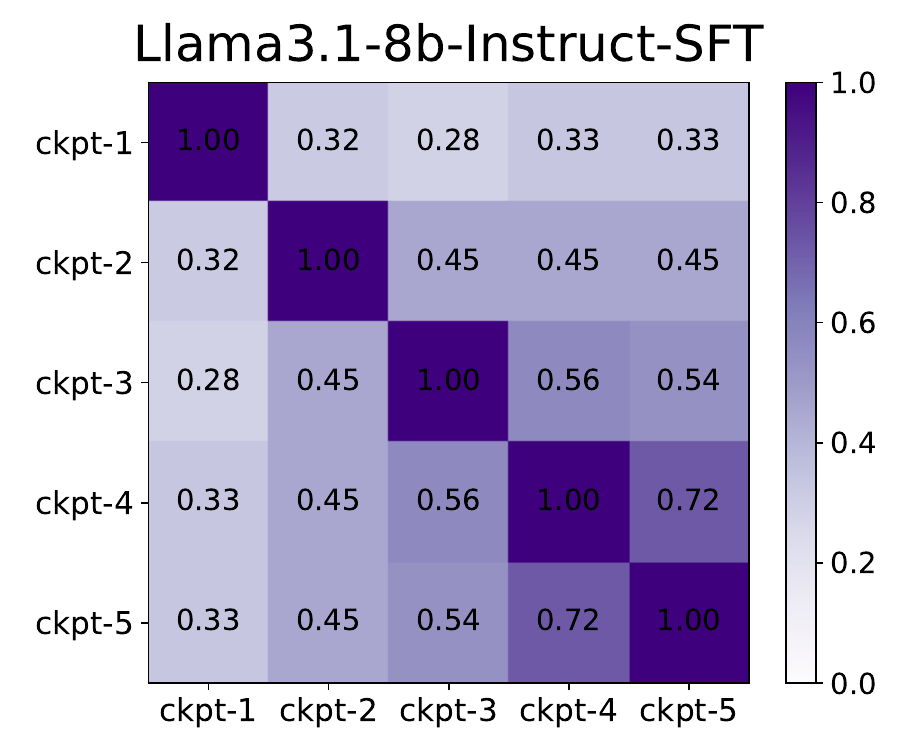}
    \hspace{-2.5mm}
    \includegraphics[width=0.5\linewidth]{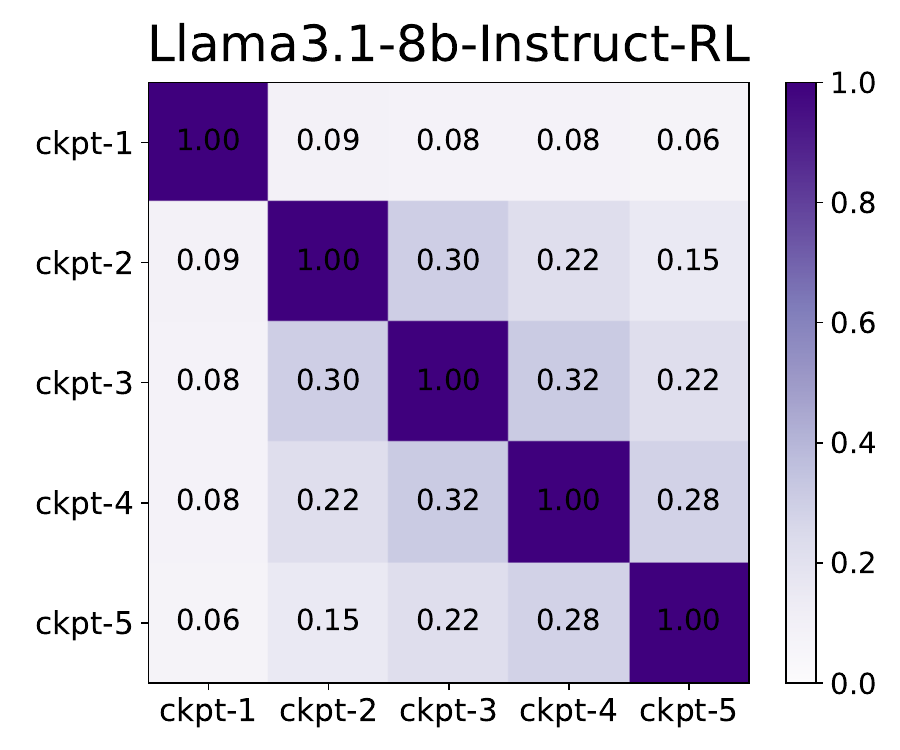}
    \caption{Feature overlap heatmaps across training checkpoints under different training paradigms on Llama3.1-8B-Instruct.}
    \label{fig:llama_overlap}
\end{figure}

\begin{figure*}[t]
    \centering
    \includegraphics[width=\textwidth]{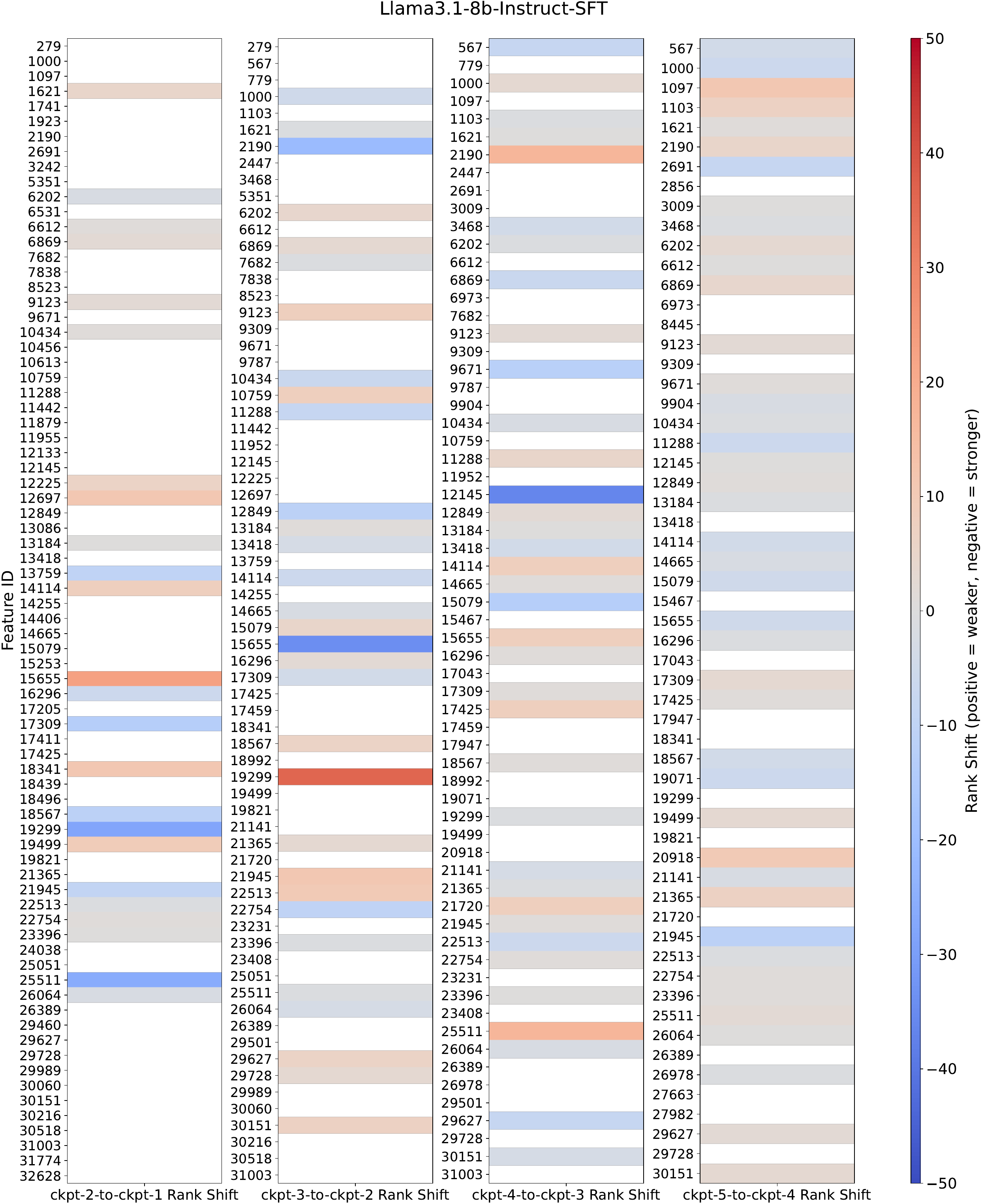}
    \caption{Feature rank shift across adjacent checkpoints during SFT on Llama3.1-8B-Instruct.}
    \label{fig:rank_shift_8b_sft}
\end{figure*}

\begin{figure*}[t]
    \centering
    \includegraphics[width=\textwidth]{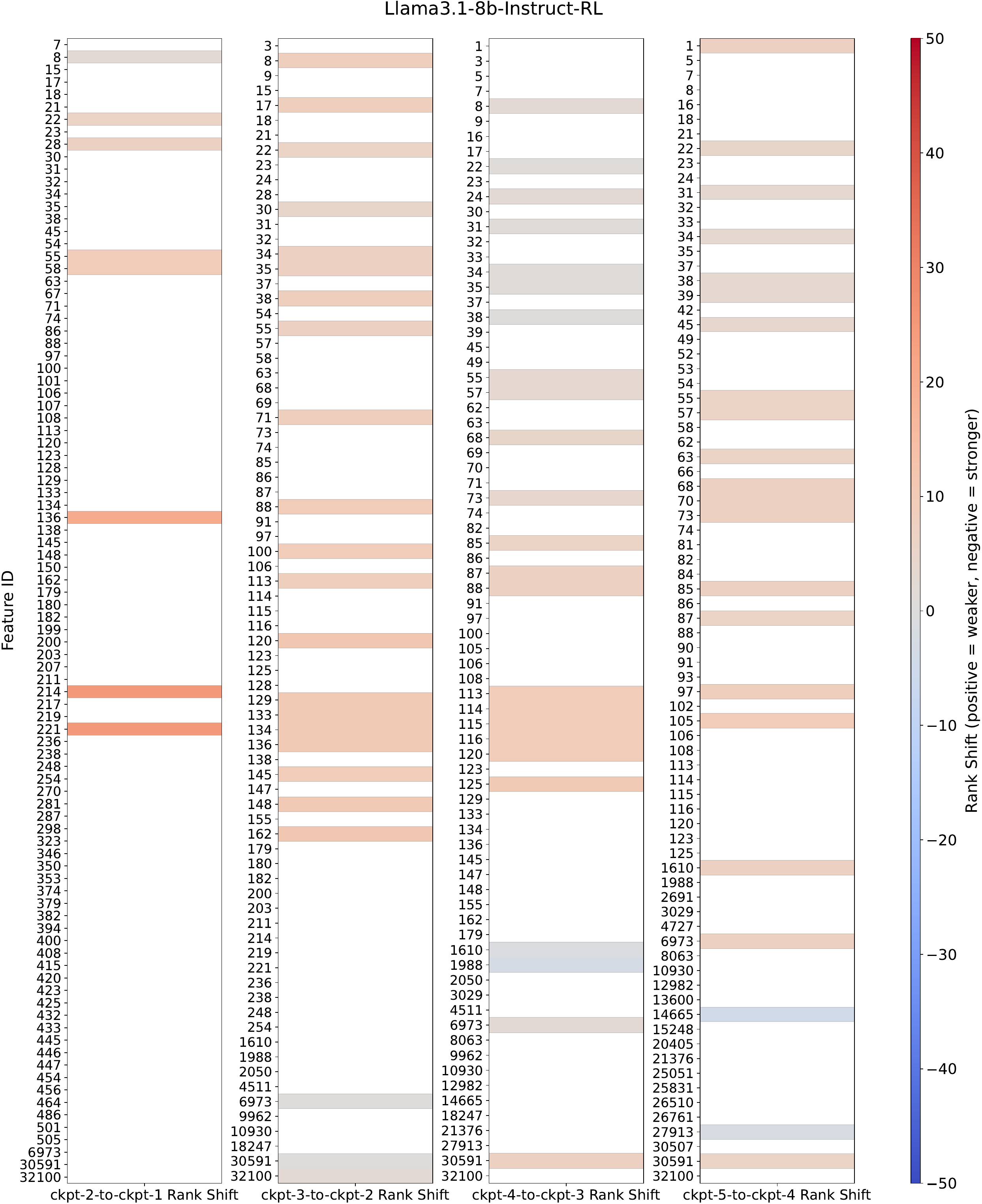}
    \caption{Feature rank shift across adjacent checkpoints during RL on Llama3.1-8B-Instruct.}
    \label{fig:rank_shift_8b_rl}
\end{figure*}

The results show consistent trends with those observed in the Qwen models. As shown in Figure \ref{fig:llama_NRN}, SFT introduces a larger number of highly model-specific features, while RL preserves the base representations and induces more restrained feature changes. In addition, Figure \ref{fig:llama_overlap}, \ref{fig:rank_shift_8b_sft}, and \ref{fig:rank_shift_8b_rl} show that SFT features stabilize early during training, whereas RL exhibits more gradual feature evolution.

\subsection{Additional MAS Results for Qwen2.5-7B}
\label{app:mas_7b}
Figure~\ref{fig:MAS_7b} provides additional MAS distributions on Qwen2.5-7B. The same qualitative trends persist: SFT induces a heavily right-skewed MAS distribution with many highly attributed features, whereas RL results in a significantly flatter distribution with fewer strongly model-specific features.

\subsection{Generalization to Unseen Tasks.}
\label{app:unseen_tasks}

To further evaluate whether the identified generalization features extend beyond the tasks used for feature identification, we conduct additional experiments on two unseen benchmarks: LogiQA \cite{liu2021logiqa}, a logical reasoning benchmark, and PIQA \cite{bisk2020piqa}, a physical commonsense question answering benchmark. These tasks differ substantially from the original evaluation set in domain and question structure, and are not used during feature selection.

We perform the same feature-level interventions as in the main experiments. Specifically, we (i) set the identified generalization features to zero in the RL-trained models, and (ii) amplify the same features in the base models, and evaluate performance on generalization-critical samples.

Table~\ref{tab:zero_unseen} shows that zeroing the generalization features in RL-trained models leads to clear performance degradation on both LogiQA and PIQA. Conversely, as shown in Table~\ref{tab:amplify_unseen}, amplifying these features in the base models consistently improves performance on the unseen tasks. These results further support that the identified features implement a general-purpose generalization mechanism rather than task-specific heuristics.

\begin{table}[t]
  \centering
  \setlength{\tabcolsep}{8pt} 
    \begin{tabular}{lcc}
    \toprule
    \textbf{Model} & \textbf{LogiQA} & \textbf{PIQA} \\
    \midrule
    Qwen3-4B-RL & -24.5  & -17.6 \\
    Qwen2.5-7B-RL & -24.0  & -11.8  \\
    \bottomrule
    \end{tabular}%
    \caption{Performance degradation on unseen tasks by zeroing generalization features in the RL-tuned model.}
  \label{tab:zero_unseen}%
\end{table}%

\begin{table}[t]
  \centering
  \setlength{\tabcolsep}{8pt}
    \begin{tabular}{lcc}
    \toprule
    \textbf{Model} & \textbf{LogiQA} & \textbf{PIQA} \\
    \midrule
    Qwen3-4B-Base & +23.3  & +28.2 \\
    Qwen2.5-7B & +24.0  & +32.9  \\
    \bottomrule
    \end{tabular}%
    \caption{Performance improvements on unseen tasks by amplifying generalization features in the base model.}
  \label{tab:amplify_unseen}%
\end{table}%


\end{document}